\documentclass[12pt,a4paper]{article}

\usepackage[utf8]{inputenc}
\usepackage[T1]{fontenc}
\usepackage{amsmath,amssymb}
\usepackage{graphicx}
\usepackage{hyperref}
\usepackage{geometry}
\usepackage{breakcites}
\title{The Role of Accuracy and Validation Effectiveness in Conversational Business Analytics}
\author{Adem Alparslan\\
\small Department of Business Analytics, FOM University of Applied Sciences \\
\small \texttt{adem.alparslan@fom.de}}
\date{}
\begin{document}

\maketitle

\begin{abstract}
This study examines conversational business analytics, an approach that utilizes AI to address the technical competency gaps that hinder end users from effectively using traditional self-service analytics. By facilitating natural language interactions, conversational business analytics aims to empower end users to independently retrieve data and generate insights. The analysis focuses on Text-to-SQL as a representative technology for translating natural language requests into SQL statements. Developing theoretical models grounded in expected utility theory, this study identifies the conditions under which conversational business analytics, through partial or full support, can outperform delegation to human experts. The results indicate that partial support, focusing solely on information generation by AI, is viable when the accuracy of AI-generated SQL queries leads to a profit that surpasses the performance of a human expert. In contrast, full support includes not only information generation but also validation through explanations provided by the AI, and requires sufficiently high validation effectiveness to be reliable. However, user-based validation presents challenges, such as misjudgment and rejection of valid SQL queries, which may limit the effectiveness of conversational business analytics. These challenges underscore the need for robust validation mechanisms, including improved user support, automated processes, and methods for assessing quality independent of the technical competency of end users.
\end{abstract}

\maketitle

\section{Introduction}
Business analytics aims to generate actionable insights that support data-driven decision-making across diverse organizational contexts. Self-service analytics, a significant development in this domain, empowers end users to independently fulfill their information needs without relying on experts such as data engineers or data scientists. By providing tools for retrieving, preparing, analyzing, and visualizing data, self-service analytics enhances flexibility and agility in addressing dynamic business demands.

Despite these benefits, self-service analytics has several limitations. While end users are often domain experts, they often lack the technical skills required for advanced analytics tasks, such as navigating complex data structures, writing code, or using machine learning techniques. This skill gap can lead to errors and continued reliance on technical experts, thus undermining the autonomy and effectiveness of self-service analytics.

Recent advances in generative AI, particularly the development of large language models, provide a transformative solution to these challenges. These models enable natural language interaction with business analytics systems, eliminating the need for technical expertise. Building on this foundation, conversational business analytics is emerging as an innovative paradigm that redefines how users interact with business analytics systems. By leveraging large language models, conversational business analytics enables users to delegate tasks such as data retrieval, analysis, and visualization to AI, which understands and generates natural language output. This approach bridges the skill gap and extends access to sophisticated analytics tools to a wider range of organizational roles.

There is a growing number of initiatives to extend self-service analytics using AI-powered natural language interface. In response, software vendors are expanding their product portfolios to leverage individual data analytics. These solutions are designed to enable end users to interact seamlessly with data, generate reports and perform a wide range of analytical tasks independently.

This study develops theoretical models, based on the expected utility theory, to identify the conditions under which conversational business analytics using Text-to-SQL outperforms delegation to human experts. Central to this analysis is the interplay between accuracy (the ability of AI to generate correct information) and validation effectiveness (the performance in correctly distinguishing between correct and incorrect information). The models examine two levels of AI support: partial support, in which the AI generates information without additional validation, and full support, which includes a validation process to increase trustworthiness. The conditions under which conversational business analytics surpass delegation to human experts are identified, with a particular focus on scenarios where validation should be performed to enhance trustworthiness and those where validation should be omitted to avoid degradation in information quality. These findings provide a structured framework for determining when and how AI-driven insight generation combined with validation can increase benefits while maintaining reliability, and offer practical guidance for implementing conversational business analytics.

The structure of this study is as follows. Chapter 2 provides an overview of traditional self-service analytics, highlighting its limitations. Chapter 3 introduces conversational business analytics and explores its transformative potential with a focus on large language models. This section also illustrates the challenges of information generation and subsequent validation, using Text-to-SQL as an example. Chapter 4 presents models that explore the dynamics of partial and full support scenarios, focusing on their implications for  information generation and validation. Finally, Chapter 5 concludes with a summary of the key findings, an acknowledgement of the limitations of the study, and a discussion of directions for future research.

\section{Traditional Self-Service Analytics}

\begin{figure}[h]
	\begin{center}
		    \includegraphics[width=0.4\textwidth]{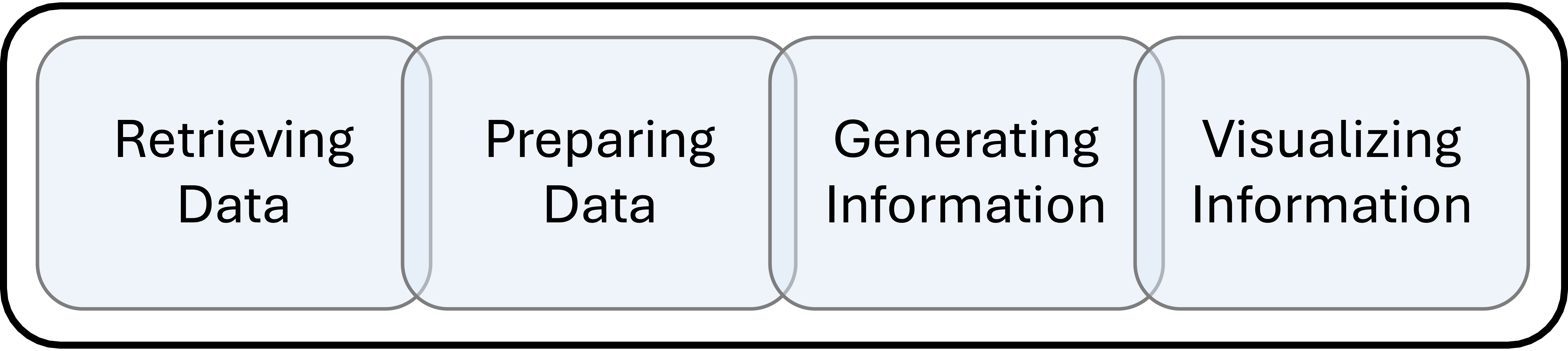} 
    \caption{Core insight generation tasks of business analytics.}
    \label{fig:tasks}
	\end{center}
\end{figure}

Business analytics focuses on generating actionable insights to support data-driven decision-making in business \cite{Chen2012, Delen2018, Holsapple2014, Vercellis2009}. To achieve these goals, business analytics systems consist of components \cite{Inmon2015, Watson2019, Prabhu2019, Nambiar2022} that automate four core tasks related to insight generation: data retrieval, data preparation, information generation and information visualization (see Fig. \ref{fig:tasks}). In addition to these core tasks, there are complementary tasks such as ensuring data and information quality and establishing governance mechanisms. Although these complementary tasks are important for achieving goals associated with business analytics, they are not discussed further in this study.

Retrieving data involves collecting raw data from various source systems, such as structured data from relational databases, semi-structured data such as JSON or XML, and unstructured data such as text or images. The next step, data preparation, includes cleansing, transformation and enrichment processes to ensure quality and usability. The prepared data are stored in data warehouses and data marts for structured analysis, whereas semi-structured or unstructured data are stored in data lakes \cite{Llave2018} or variants, prepared on demand.

Information generation applies the three main methods of business analytics \cite{Delen2013}: descriptive, predictive, and prescriptive analytics. Descriptive analytics summarizes historical and real-time data, providing insights into past and current performance using key performance indicators (KPIs), trend analysis, and target-performance comparisons \cite{Delen2018}. Predictive analytics uses statistical models and machine learning techniques to forecast future developments, such as predicting customer churn or market trends. Prescriptive analytics recommends actions based on predictions, utilizing planning, simulation, and optimization methods to identify effective decisions. These methods are complementary, with prescriptive analytics relying on both the descriptive and predictive results.

Finally, visualizing information transforms analytical results into intuitive formats that support decision-making \cite{Stodder2015}. Effective visualizations, such as dashboards simplify complex insights, enable the exploration of trends, identify opportunities, and act quickly. 

Day-to-day information needs are typically met through predefined reports provided by business analytics systems.  However, when new information needs arise, self-service analytics \cite{Imhoff2011, Alpar2016, Michalczyk2020} enables end users to fulfill information needs independently, providing a viable alternative to the traditional reliance on expert (e.g. data engineers and data scientists) intervention. In this context, an end user refers to any organizational member who uses information to make decisions, whether to inform their own decisions or prepare information for others, such as supervisors. These individuals are typically domain experts with extensive knowledge in their respective fields, and are distributed across various departments within the organization.

Self-service analytics offers significant benefits \cite{Imhoff2011, Alpar2016, Palys2023}, enabling flexible and timely information generation to support the early identification of opportunities and risks. This allows for faster decision making, both to capitalize on positive outcomes and to mitigate negative impacts. Self-service analytics also falls under the umbrella of ''end-user computing'' \cite{Nelson1989}, where tasks traditionally managed by specialized departments are now performed by members of the organization. By eliminating the need to delegate information retrieval tasks to experts, self-service analytics removes dependencies and delays. In addition, it reduces agency costs associated with delegation \cite{Gurbaxani1990} by minimizing exposure to hidden actions, which in turn reduces the need for incentives or oversight.

Self-service analytics for descriptive purposes have traditionally relied on the data mart approach, enabling users to access predefined information and perform OLAP operations \cite{Codd1993, Chaudhuri1997}, such as drill-down and roll-up. However, data -- customer, product, or process-related -- often exist outside of data marts, in data warehouses, data lakes, or source systems.  Although real-time OLAP and in-memory computing have improved responsiveness, generating actionable insights often requires expert assistance, leading to delays. To address this, greater flexibility is needed, allowing the independent retrieval and preparation of data, performing analytics, and visualizing results \cite{Alpar2016}. This autonomy enhances decision-making and supports agile responses to changing business needs.

The effective use of self-service analytics requires technical expertise \cite{Schmarzo2023}, which many end users lack. While end users typically have domain-specific knowledge, they often lack skills in critical areas such as data retrieval, modeling, machine learning, programming and navigating complex data structures \cite{Passlick2023}. For example, Microsoft's financial data warehouse, comprising 632 tables, over 4,000 columns and 200 views \cite{Floratou2024}, illustrates the significant challenges of navigating large and complex databases. This knowledge gap not only increases the likelihood of errors, but also limits the ability to independently generate actionable insights.

\begin{figure}[h!]
    \centering
    \includegraphics[width=0.4\textwidth]{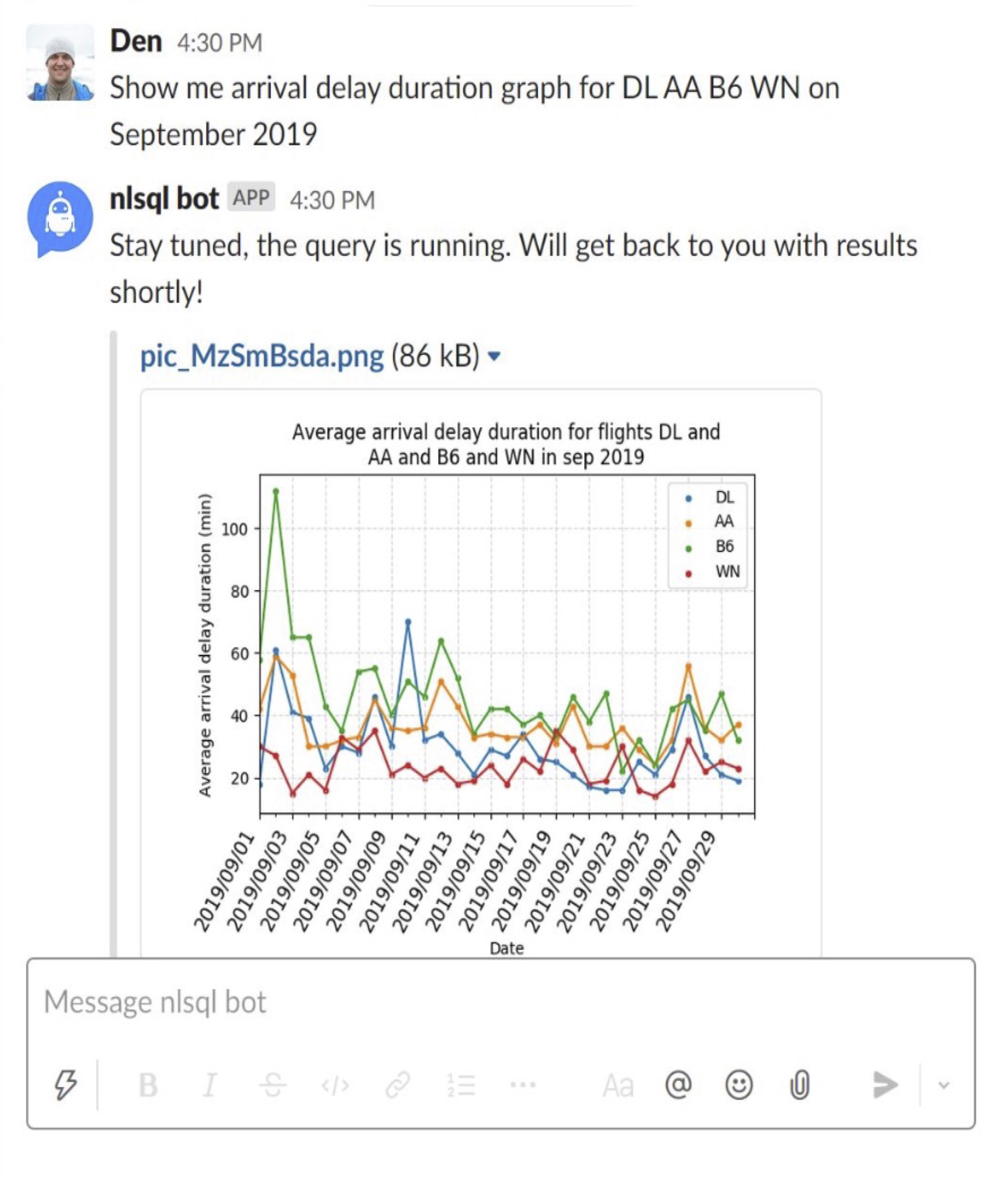}
    \caption{CBA facilitating natural language requests and generating insights \cite{Alparslan2024_b}}
    \label{fig:CBA_1}
\end{figure}

Several studies underscore the importance of technical skills in the effective application of self-service analytics \cite{Imhoff2011, Lennerholt2021, BanHani2019}. Imhoff and White highlight the necessity of user-friendly tools, noting that ''sophisticated analytics are often too intricate, complex, or difficult to construct for many information workers.'' Similarly, Lennerholt, Laere, and Söderström emphasize the dual importance of intuitive tool design and comprehensive training programs to address the technical knowledge gap among non-technical users \cite{Lennerholt2021}. 
In an empirical study, Alparslan and Hügens examine the challenges faced by small and medium-sized enterprises in using business analytics systems \cite{Alparslan2023}. Their findings show that for around 70\% of respondents, inadequate technical skills are a key barrier to turning raw data into actionable insights.

\section{Conversational Business Analytics}
\subsection{Overview}

Business analytics is undergoing significant development with the emergence of generative AI, giving rise to a new approach termed “Conversational Business Analytics” (CBA). This new paradigm leverages natural language processing to address persistent challenges in traditional self-service analytics, particularly the technical skill gap among end users. CBA facilitates natural language interactions for core business analytics tasks, significantly improving their accessibility and effectivity.

CBA shifts the focus from traditional graphical user interfaces to natural language-driven interactions, supported by advancements in large language models (LLMs) \cite{Vaswani2017, Devlin2019, Brown2020}. These models process natural language by tokenizing text and employing self-attention mechanisms to interpret the contextual relationships. LLMs enable context-aware and relevant outputs by calculating the most probable response token by token. Trained on extensive text corpora, LLMs optimize billions of parameters to achieve high performance in natural language understanding and generation. As depicted in Fig. \ref{fig:CBA_1}, CBA facilitates the transformation of natural language inputs into actionable outputs such as structured reports or visualizations.

CBA supports structured data (e.g., from data warehouses), semi-structured data (e.g., from data lakes), and unstructured data (e.g., textual documents). For structured and semi-structured data, semantic parsing techniques are used to convert natural language requests into executable code. For unstructured data, LLMs extract insights through advanced text processing methods. Table \ref{tab:cba_tasks} summarizes the core tasks and their corresponding references for recent advancements.

\begin{table}
\caption{\textbf{Selected advancements in CBA regarding the core tasks of business analytics.}}
\label{tab:cba_tasks}
\setlength{\tabcolsep}{3pt}
\begin{tabular}{p{250pt}p{150pt}}
\hline
Task & References \\
\hline

Data retrieval & \cite{Tran2024}, \cite{Thapa2024}, \cite{Katsogiannis2023}, \cite{Jeong2023}, \cite{Guo_RAG_2023}  \\
	Data Preparation & \cite{Zhang2023}, \cite{Sharma2023}, \cite{Koch2023}, \cite{Chen2024}, \cite{Nasseri2023} \\
	Information extraction \& generation & \cite{Musazade2024}, \cite{Arasteh2024}, \cite{Li2023} \\
	Information visualization &  \cite{Wu2024}, \cite{Stockl2024}, \cite{Maddigan2023}, \cite{Monteiro2023}, \cite{Dejeu2024} \\
\hline
\end{tabular}
\end{table}

CBA also introduces interactive exchanges between users and AI, enabling refinement of requests, validation of outputs, and clarification of insights. Such interaction underscores its user-centric and adaptive design, as exemplified by tools such as OpenAI’s ChatGPT Advanced Data Analysis, which guides users through tasks like training machine learning models and subsequent model optimization \cite{Arasteh2024}.

\begin{figure*} [h!]
    \centering
    \includegraphics[width=0.7\textwidth]{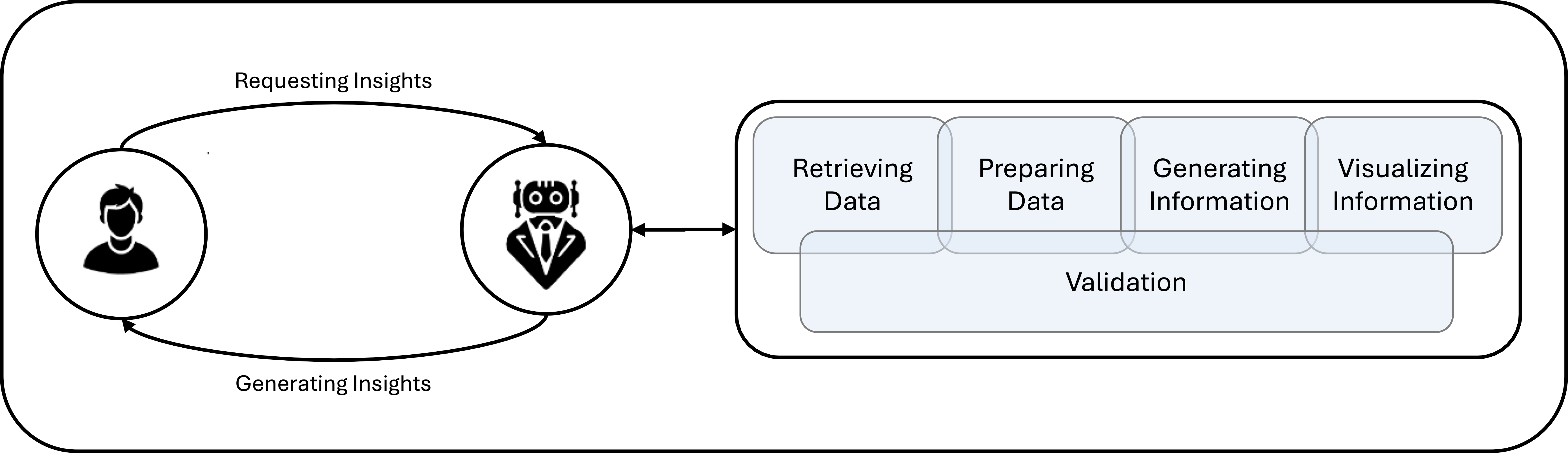} 
    \caption{Delegation of core tasks to AI.}
    \label{fig:delegation_AI}
\end{figure*}

By integrating natural language processing, interactive communication, and support for diverse tasks and data types, CBA introduces a novel form of delegation in business analytics (see Fig.  \ref{fig:delegation_AI}). Through CBA, complex tasks, such as data processing and information generation, can be assigned to AI, enabling autonomous generation and validation of insights on behalf of the end user. This delegation has the potential to reduce dependence on human experts, while addressing the limitations associated with traditional self-service analytics.

Given the dynamic nature of CBA, this study focuses on Text-to-SQL \cite{Katsogiannis2023, Beck2024, Askari2024, Elgohary2020, Yao2019} as a key semantic parsing technology. Text-to-SQL, a component of natural language interfaces to databases \cite{Li2024_NLDB}, translates natural language prompts into Structured Query Language (SQL) queries, enabling the retrieval and transformation of data from relational databases. This technology supports both data retrieval and the preparation of comprehensive workflows for generating actionable insights.

The origins of Text-to-SQL can be traced back to Codd's pioneering work in the 1970s, proposing natural language interfaces for ''casual users'' to interact with relational databases \cite{Codd1974}. Early implementations, such as rule-based systems \cite{Warren1982}, were limited in flexibility, but subsequent advancements, including Long Short-Term Memory (LSTM) networks \cite{Zhong2017}, have improved the handling of sequential inputs. More recently, transformer-based architectures have replaced LSTM networks, demonstrating superior performance in complex natural language processing tasks. Modern Text-to-SQL systems leverage LLMs for both training and inference, significantly improving accuracy and capability.

The primary focus of this study is on the accuracy of Text-to-SQL, which is defined as its effectiveness in generating correct SQL queries. Ensuring accuracy is crucial, because errors in query generation can lead to incorrect insights and suboptimal decisions. Additionally, this study examines validation effectiveness, which reflects the ability to correctly distinguish between correct and incorrect information. Validation effectiveness plays a key role in verifying the reliability of the generated information, particularly in high-stake scenarios where flawed data can have significant consequences. The dual challenges of ensuring the effectiveness of information generation and validation are explored in detail in the following sections.

\subsection{Accuracy}
Delegating tasks to human experts is grounded in their ability to correctly interpret and respond to information needs as well as to build a shared understanding of the underlying goals and requirements \cite{Glinz2015, Iriarte2020}. A similar dynamic exists when interacting with AI. The AI must understand the user's request to generate the desired outcome. However, while natural language is flexible and often ambiguous, SQL is highly structured and formal. For instance, the key figure ''material availability'' may be interpreted differently by the logistics and maintenance departments within the same organization because of the coexistence of multiple terminological systems \cite{Mori1995, Westbrook1992, Hirschheim1995}. Beyond terminological differences, ambiguities may arise from the linguistic complexity of the request itself, such as vague formulations. To translate such requests into correct SQL queries, AI must map the terms used in the request to the corresponding tables and columns within the data model (schema mapping). This requires a deep understanding of the semantics of the data model, including the specific meaning of the fields and the relationships between tables. Thus, the AI must not only understand the request but also identify the relevant tables and columns that match the user's inquiry.

The AI's ability to interpret a natural language request and translate it into the correct SQL code is referred to as its accuracy. Various definitions of accuracy can be found in the literature, reflecting different perspectives and contexts \cite{Katsogiannis2023, Yu2019, Tomova2023, Jeong2023}. Syntactic accuracy examines whether the SQL statements generated by AI are executable. The execution accuracy evaluates whether the SQL query generated by the AI produces the expected result, even if the query's structure differs from a reference SQL statement (often referred to as the ''gold standard''). In contrast, exact match accuracy is more stringent, as it requires not only the correct result but also that the generated SQL query exactly matches the gold standard.  

Two main strategies have been proposed to enhance the accuracy of Text-to-SQL: fine-tuning and prompt design. Each of these methods addresses different aspects: fine-tuning focuses on adjusting the model parameters of a LLMs, whereas prompt design involves crafting the input to the LLM.

\begin{enumerate}

\item Fine-tuning:  
Training an LLM from scratch is resource-intensive and requires vast amounts of data and computational power \cite{Raffel2020, Pourreza2024, Thorpe2024, Guo_RAG_2023}. As an alternative, organizations can leverage transfer learning, wherein a pre-trained LLM that has already learned general language representations is refined with domain-specific data. Fine-tuning involves optimizing the parameters of a pre-trained LLM to enhance its performance on specific tasks and ensure better alignment with particular organizational needs and domain-specific requirements. In the context of Text-to-SQL, fine-tuning involves training the model on labeled datasets consisting of pairs of natural language requests and their corresponding SQL statements. This process helps the model to understand the organization's specific data models, schema, and terminology, enabling it to generate correct SQL queries based on user input.

A major drawback of fine-tuning is the risk of overfitting. This occurs when the model becomes overly specialized in domain-specific data and is less capable of adapting to other requests, making it more difficult to generalize to different contexts. Moreover, ongoing maintenance poses challenges: as data repositories (such as data warehouses or data marts) evolve, schemas frequently change. Therefore, the fine-tuned LLM may require periodic updates to remain effective, increasing operational complexity and costs. Fine-tuning is not a one-time task; it requires continuous updates to avoid becoming outdated as the organizational data landscape evolves.

\begin{figure*}
    \centering
    \includegraphics[width=0.7\textwidth]{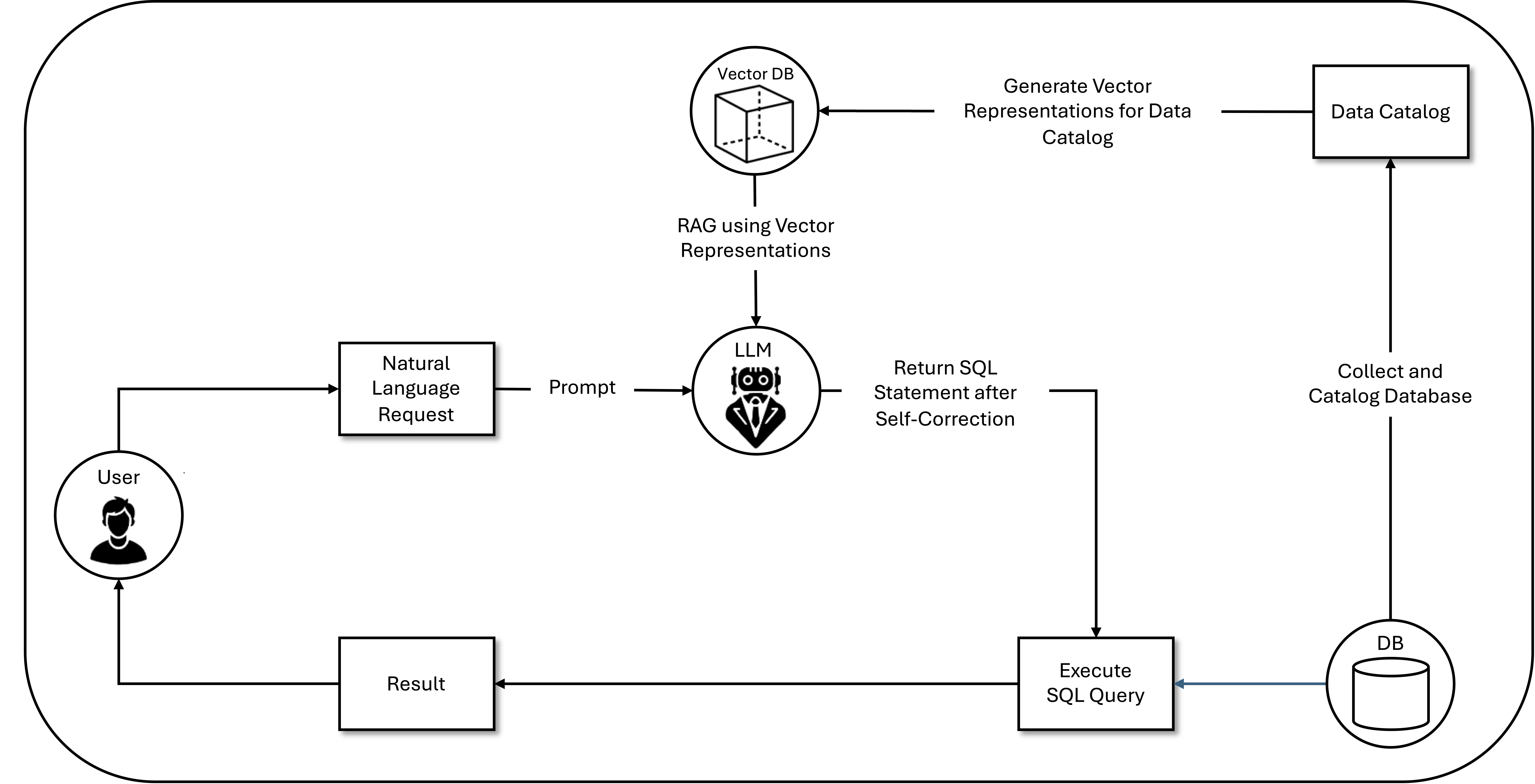}
    \caption{Integration of LLM and Retrieval-Augmented Generation for Text-to-SQL; based on \cite{Guo_RAG_2023, Eusebius2024, Beck2024, Thorpe2024}.}
    \label{fig:rag}
\end{figure*}

\item Prompt Design:  
Unlike fine-tuning, prompt design \cite{Guo_RAG_2023, Liu2023, Pourreza2024} does not alter the underlying parameters of the LLM. Instead, it leverages the model's generalization capabilities by crafting prompts that guide the LLM toward correct SQL query generation. In few-shot learning \cite{Brown2020, Liu2023}, for example, a user's request is enriched with additional contextual information, such as data model details or examples of analogous SQL queries. The LLM processes the enriched prompt to infer the appropriate SQL query. Here, the LLM does not acquire new knowledge; rather, it uses its existing understanding along with the provided context to generate SQL queries that it has not previously encountered. Retrieval-Augmented Generation (RAG) further enhances prompt design by retrieving relevant contextual information, such as metadata stored in a vector database. Upon receiving user input, RAG integrates this context into the prompt, enabling the LLM to generate SQL queries. These queries undergo automated checks for syntax, schema alignment, and compliance with user requirements. If discrepancies are detected, the LLM’s self-correction mechanism refines the query. Once validated, the query is executed, and the information is presented to the user. Fig. \ref{fig:rag} illustrates the integration of LLMs with RAG, highlighting seamless context retrieval and query refinement.

Prompt design has the advantage of eliminating the need for continuous fine-tuning, saving both time and computational resources. However, as more contextual information is added, the number of tokens increases, pushing against the limits of the LLM's context window  \cite{Chen2023_window} -- the maximum number of tokens it can process in a single instance. Although modern LLMs have extended context windows, this limitation can still lead to information loss in very complex or lengthy prompts, potentially impacting the overall accuracy of the generated query.
\end{enumerate}

The integration of LLMs has significantly improved Text-to-SQL performance. On the SPIDER benchmark \cite{Yu2019}, which assesses the generalization capabilities of Text-to-SQL models across a wide variety of databases, the performance of leading models has markedly improved, with the execution accuracy increasing from approximately 54\% to 91\%. The exact match accuracy, which was initially approximately 5\%, was increased to 82\%. In comparison, the BIRD benchmark \cite{Li_BIRD_2023}, which presents even greater complexity, reveals that the current LLM-based Text-to-SQL models achieve execution accuracy of around 73\%, underscoring the challenges posed by this benchmark.

\subsection{Validation Effectiveness}  
When delegating information production to human experts, there is a challenge of ''hidden actions'' where the actions of experts are not fully observable or assessable, potentially leading to misalignment of interests \cite{Gurbaxani1990}. Similarly, when delegated to AI, the issue shifts to the transparency and reliability of outputs. While human experts may act in self interest, leading to agency costs through false information or additional incentives, AI systems present the challenge of output opacity. End users cannot evaluate the correctness of SQL queries and the resulting KPIs until after the implementation, risking decisions based on false information with economic consequences. Ensuring the reliability of AI-generated information is crucial, particularly in high-stakes decision-making scenarios where false information can result in significant negative business outcomes. The requirement for reliability is further intensified by the inherent tendency of LLMs to produce hallucinations \cite{Lin2024, Perkovic2024}, resulting in fabricated or erroneous SQL queries that may appear credible. Consequently, robust techniques are essential for validating the generated SQL queries. 

Techniques for validating automatically generated SQL queries fall within the broader field of explainable AI \cite{DARPA2016, Rudin2019, Minh2022, Langer2021, Gunning2019}. Since the advent of expert systems, researchers have emphasized the pivotal role of explanations in enhancing the interpretability of AI-generated results \cite{Wallis1982, Nakatsu2006}. In recent years, the growing demand for transparency in machine learning has prompted significant advances in the field of explainable AI, leading to an expansion of both the scope and sophistication of explanation techniques. The primary objective of explainable AI is to provide explanations that elucidate the underlying processes behind AI outputs, thereby facilitating tasks such as debugging, regulatory compliance, the establishment of user trust, and the effective utilization of AI.

In the specific context of Text-to-SQL, validation necessitates the provision of explanations that help to comprehend and verify AI-generated SQL queries. These explanations must detail the structure of the queries, including the tables, fields, and operations (e.g., filtering, grouping, or joining) involved. Such detailed information enables users to critically evaluate the logical coherence of a query and its alignment with their intended objectives, thereby supporting an independent validation process. Effective validation ensures that users can reliably differentiate between the correct and incorrect SQL queries. However, validation is ineffective if incorrect SQL queries are misclassified as correct, or if correct queries are misclassified as incorrect.

Validation in Text-to-SQL extends beyond mere explanation and often encompasses techniques for error correction. These techniques reflect the interactive nature of CBA, and aim to support users in identifying and rectifying errors in SQL queries. In this study, the interpretability of the AI outputs and their correction are treated separately, with a focus on validation effectiveness as the primary metric. Validation serves as the foundation for reliable decision-making, as correct validation is essential for subsequent error correction.  

\begin{figure*}[h]
    \centering
    \includegraphics[width=0.9\textwidth]{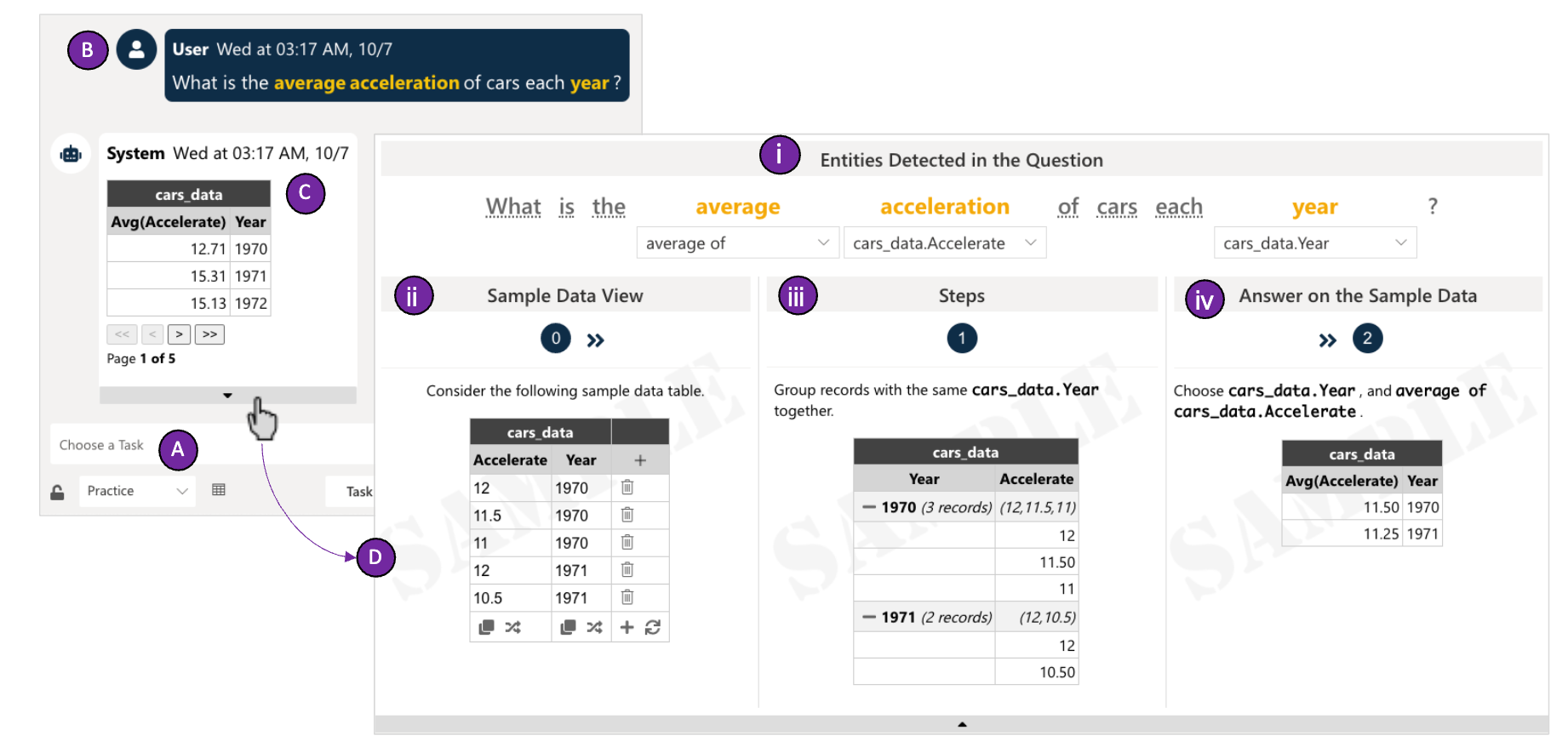} 
    \caption{Decomposition technique for explaining SQL queries \cite{Arpit2021}.}
    \label{fig:decomposition}
\end{figure*}

The current literature identifies three techniques for validating and explaining SQL queries (see \cite{Zhang2023__} for an overview). Decomposition techniques (see Fig. \ref{fig:decomposition}) break SQL queries into components such as tables, fields, and operations, presenting intermediate results based on sample data to trace each element's contribution to the final output \cite{Arpit2021}. Visualization techniques use graphical representations to map relationships among query components, thereby enhancing clarity and navigation in complex query structures \cite{Leventidis2020, Miedema2021, Murakawa2011}. Dialogue techniques employ LLMs to provide natural language explanations, enabling users to interactively explore and refine queries by addressing specific components or underlying data model \cite{Tian2024, Elgohary2020, Kokkalis2012, Gur2018, Xiaxia2021, yao2019model}. 
 
Empirical studies on validation techniques for Text-to-SQL reveal varying levels of effectiveness in supporting SQL comprehension and error correction. In a user study, Ning et al. observed that decomposition, visualization, and dialogue techniques achieved an average effectiveness rate of approximately 56\% in detecting and correcting errors \cite{Zhang2023__}. In contrast, Tian et al. demonstrated that their technique, which combines elements of decomposition and dialogue techniques, significantly outperformed existing techniques, achieving an average effectiveness in validation and error correction  of approximately 85\% \cite{Tian2024}. These findings underscore the substantial progress in user-based validation and error correction, showcasing their transformative potential to enhance the reliability of AI-generated SQL queries. 

Tian et al. showed that their technique consistently delivers effective results regardless of the technical competency of end users \cite{Tian2024}. However, the boundaries of user-based validation in complex business contexts, such as data warehouses with hundreds of interrelated tables and fields, require further research. Adding to the complexity, the explanatory information in Text-to-SQL differs significantly from the explainable AI approaches commonly used to interpret machine learning models. Explanations in machine learning are typically closely tied to business language, such as: "The model predicts that the customer will have poor creditworthiness because the customer's features exhibit these values." In contrast, the explanatory information in Text-to-SQL has a distinctly technical focus, aiming to improve the user's understanding of how the underlying SQL query operates. The expertise and experience of the end users play a critical role in their ability to comprehend the software code and, consequently, evaluate the correctness of the generated information.

Furthermore, the complexity of SQL queries, particularly those involving numerous subqueries and joins, combined with the cognitive load they impose on users, affects their ability to comprehend software code (see, for example, \cite{Munoz2020}, which references developers rather than end users). This cognitive load is further exacerbated by the architecture of the underlying data model \cite{Chan2005}, especially in cases where data models include extensive dependencies and abstract schema structures that pose significant challenges for end users.

Additionally, end users typically perform validation only occasionally when generating new SQL queries to meet a new information need. Consequently, they are unable to develop long-term knowledge of SQL or to gain a deeper understanding of the underlying data model. This challenge has also been noted in traditional self-service analytics \cite{Vujosevic2018}, where users face difficulties in building experiential knowledge of the tools and data models employed for individual information generation.

\section{Models of CBA}
\subsection{Basic Assumptions}

This study introduces theoretical models grounded in rational choice theory, with a focus on expected utility theory \cite{Neumann2007, Tunney2024}, to evaluate the effective use of CBA. The examination centers on the example of Text-to-SQL. Two primary influencing factors are considered in this study:  

\begin{itemize}
    \item accuracy, defined as the AI's ability to generate correct SQL queries.  
    \item validation effectiveness, defined as the ability to distinguish between correct and incorrect SQL queries.
\end{itemize}

It is assumed that the end user possesses knowledge of the levels of these factors, which influence how effectively AI performs in handling insight-generation tasks, and is able to assess their suitability accordingly. Delegation to a data engineer is considered as an alternative to using AI for insight generation. 

In an ideal scenario, both the accuracy and validation effectiveness would be perfect, ensuring the generation of reliable and actionable information. However, these factors are inherently probabilistic and are subject to imperfections. The models examine how the interaction between accuracy and validation effectiveness influences the effectiveness of Text-to-SQL and its potential advantages over delegation to a human expert. Specifically, the models evaluate how varying levels of these factors affect the relative advantage of delegating tasks to AI, with or without validation support. By exploring this interplay, the models aim to identify the conditions under which Text-to-SQL can serve as a viable alternative to human expertise, providing a structured framework for its adoption in CBA. 

These  models assume a risk-neutral perspective, where the end user has a linear utility function. This simplification allows for a focus on the average outcomes (expected values), while ignoring the variability in potential gains and losses. By assuming linearity, the models quantify all the values in monetary units.

The analysis begins with the end user's need for information that cannot be provided through standard reporting. The current value of a KPI is critical for informed business decisions. While the end user has access to data marts, the necessary data reside in data warehouses, requiring an SQL query for extraction. This query retrieves data from the data warehouse, prepares it, and ultimately generates the required KPI. A correctly formulated SQL query produces a valid KPI, enabling optimal decisions and yielding a net profit of \(+1\). Conversely, an erroneous query generates a misleading KPI, resulting in a suboptimal decision and a financial loss of \(-1\). The analysis assumes that the data warehouse contains high-quality, undistorted data, making the correctness of the SQL query the sole determinant of the information quality. Despite possessing expertise in business processes, the end user lacks the technical skills required to effectively apply traditional self-service analytics.

Currently, no data engineer is available to create the KPI, and such a resource will only be accessible later. The delayed delivery of the KPI results in a reduced (net) profit \(v\) (where \(0<v<1\)). This profit is lower than the achievable maximum of \(+1\), reflecting the adverse effects of the delayed action. In extreme cases, \(v\) may even approach zero, rendering KPI almost valueless. Additional factors, such as the effort required to align requirements and potential motivational challenges that could negatively impact collaboration between end user and data engineer, are not considered. Consequently, \(v\) serves as a measure of the urgency of KPI procurement, with lower values of \(v\) indicating greater urgency.

The KPI generation process can be delegated to an AI equipped with a natural language interface powered by LLMs. AI offers two levels of assistance: partial support (PS) and full support (FS). Under PS, the AI translates the natural language request into SQL code, which is executed on the data warehouse to calculate the KPI. The KPI is then used to inform decision-making, resulting in a business impact. FS includes an additional validation step to ensure the correctness of the SQL query. During this step, the end user is provided with explanations of the generated SQL query, which is designed to ensure transparency and enable the independent validation of its correctness. If the SQL query is deemed correct, the process concludes with confirmation, allowing to act on the KPI. If the query is identified as erroneous, it is rejected and no action is taken, avoiding both gain and loss. 

In contrast to the dynamic, iterative nature of CBA, this study adopts a static perspective in which KPI generation and validation occur in a single execution cycle. This idealized approach isolates the roles of information generation and validation, enabling a focused analysis of their individual effects on the effectiveness of Text-to-SQL. The static perspective was also chosen, as iterative processes for information generation and validation within CBA are still under development.

It is further assumed that the end user incurs no direct costs (e.g., for data storage, processing, or the use of LLM) when utilizing Text-to-SQL. This aligns with the prevailing practice in traditional self-service analytics, in which no fees are charged for individual instances of information generation.

Because traditional self-service analytics is not feasible owing to a lack of expertise, the end user evaluates whether the KPI  should be generated via delegation to AI (either PS or FS), or whether it is more advantageous to wait for the data engineer to receive the KPI at a later time. The end user selects the option with the highest benefit. When CBA (either PS or FS) offers a higher expected value than the profit gained through the data engineer, it is preferred. Conversely, if the expected value from CBA is lower, human delegation is preferred. In such cases, CBA is deemed ineffective, as it fails to provide timely and decision-relevant information. 

Table \ref{tab:symbols} provides the mathematical symbols used in the following and their corresponding explanations.

\begin{table}
\caption{\textbf{Mathematical symbols used and their explanations.}}
\label{tab:symbols}
\setlength{\tabcolsep}{3pt}
\begin{tabular}{p{60pt}p{360pt}}
\hline
Symbol & Explanation \\
\hline

$\alpha$ & accuracy  \\
	$\beta$ & validation effectiveness \\
	$v$ & profit obtainable through data engineer \\
	$E_{PS}(\alpha)$ & expected value in PS \\
	$E_{FS}(\alpha, \beta)$ & expected value in FS \\
	$\alpha_{PS}^*$ & accuracy threshold for PS to outperform data engineer \\
	$\alpha_{FS}^*$ & accuracy threshold for FS to outperform data engineer \\
	$\beta^*$ & threshold for validation effectiveness  to outperform data engineer \\
	$\beta^{**}$ & threshold for validation effectiveness to outperform PS \\
\hline
\end{tabular}
\label{tab1}
\end{table}
 
\subsection{Partial Support (PS)}

The first stage involves the end user evaluating the feasibility of PS. PS addresses the competency gaps that prevent end users from utilizing traditional self-service analytics by automating the translation of natural language inputs into SQL queries. This capability eliminates the need for technical proficiency in SQL syntax, allowing users to focus on decision-making rather than query formulation.

The success of PS depends primarily on the accuracy of AI in translating natural language requests into correct SQL queries. As outlined earlier, current advancements, such as fine-tuning combined with prompt design, are employed to enhance AI's information generation capabilities. Given the probabilistic nature of LLMs, the generation of SQL queries is treated as a discrete random variable. The probability of generating a correct SQL query, denoted as \(\alpha\) (where \(0 < \alpha < 1\)), corresponds to execution accuracy. This probability is determined by the ratio of successfully generated SQL queries to the total number of natural language requests of a similar type.

The expected value \(E_{PS}(\alpha)\) under PS is calculated as the weighted sum of possible outcomes. A correct SQL query yields a profit of \(+1\), whereas an incorrect SQL query results in a loss of \(-1\). These outcomes are weighted by the probabilities of success (\(\alpha\)) and failure (\(1-\alpha\)), respectively. Therefore, the expected value is expressed as follows:

\[
E_{PS}(\alpha) = \alpha \cdot (+1) + (1-\alpha) \cdot (-1) = 2\alpha - 1.
\]

Delegation to AI through PS is preferred over waiting for a data engineer when the expected value \(E_{PS}(\alpha)\) exceeds the delayed profit (\(v\)). This condition can be formalized as the ''AI delegation condition of PS'':

\begin{equation}
	\begin{aligned}
E_{PS}(\alpha) > v \Leftrightarrow \alpha^*_{PS} > (1+v)/2.		
	\end{aligned}
	\label{eq:cond_1}
\end{equation}

Condition \eqref{eq:cond_1} defines the threshold of accuracy that must be exceeded for PS to be more advantageous than relying on a data engineer. When this condition is met (\(\alpha > \alpha^*_{PS}\)), PS is viable and enables effective KPI generation without relying on human expertise. This is particularly beneficial since traditional self-service analytics is constrained by the end user’s limited technical skills. Conversely, if this condition is not satisfied (\(\alpha \leq \alpha^*_{PS}\)), the expected value from PS falls below the profit achievable through delegation to a data engineer, rendering PS unsuitable for time-sensitive, decision-critical business analytics.

\subsection{Full Support (FS)}

In the second phase, the AI-generated SQL query is reviewed by the end user to determine whether a correct KPI has been generated, ensuring transparency about the functionality of the SQL query. Techniques, such as decomposition, visualization, and dialogue-based interaction, are currently being explored to support users in comprehending SQL queries. However, user-based validation entails uncertainty. Both the user’s comprehension and the complexity of the SQL query and its underlying data model significantly influence the success of validation. Additionally, inherent uncertainty arises in dialogue-based interactions because AI-generated explanations are based on probabilities rather than deterministic rules.

To address this uncertainty, the validation effectiveness is modeled as a random variable denoted as \(\beta\) (where \(0 < \beta < 1\)). This probability reflects the likelihood that a user can correctly assess the SQL query based on the support provided. It is defined as the ratio of successful validations to the total number of validation attempts. By adopting this probabilistic framework, the model accounts for the inherent uncertainties associated with validation processes.

The end user takes action only if an SQL query is classified as correct. The probability of correctly classifying a correct SQL query is \(\alpha \cdot \beta\), resulting in a gain of \(+1\). Conversely, the probability of incorrectly classifying an incorrect SQL query as correct is \((1-\alpha) \cdot (1-\beta)\), which results in a loss of \(-1\). However, if a correct SQL query is mistakenly classified as incorrect or an incorrect SQL query is correctly classified as incorrect, no action is taken, and neither a gain nor a loss occurs.

Thus, the end user's expected value \(E_{FS}(\alpha, \beta)\) when FS is provided is given by

\begin{equation*} 
	\begin{aligned}
		E_{FS}(\alpha, \beta) & = \alpha \beta \cdot (+1)  
		+ (1-\alpha)(1-\beta) \cdot (-1) \\
		& = \alpha + \beta - 1.
	\end{aligned}
\end{equation*}

The expected value \(E_{FS}(\alpha, \beta)\) is influenced by two key factors: accuracy and validation effectiveness. Higher values of \(\alpha\) indicate improved accuracy in generating correct SQL queries, whereas higher values of \(\beta\) represent enhanced effectiveness in validating the correctness of those queries. The combined improvement in these factors increases the expected value of FS, making it more likely to yield reliable and actionable KPI. This relationship highlights the complementary role of accuracy and validation effectiveness in ensuring the success of FS in CBA.

\subsection{Comparison of Full Support with Partial  Support and Delegation to the Data Engineer}

The end user will choose FS only if it provides a higher expected value than the profit achievable through delegation, 
\[
E_{FS}(\alpha, \beta) > v,
\]
and offers a better outcome than PS, 
\[
E_{FS}(\alpha, \beta) > E_{PS}(\alpha). 
\]
For FS to outperform delegation to a data engineer, the following condition, referred to as the ''AI delegation condition for FS'' must be satisfied:
\begin{equation}
\begin{aligned}
E_{FS}(\alpha, \beta) & > v  \Leftrightarrow \beta^*  > (1 - \alpha) + v.
\end{aligned}
\label{eq:cond_2}
\end{equation}
Condition \eqref{eq:cond_2} requires the validation to be sufficiently effective and robust (e.g., against the shortcomings of the end user conducting the validation based on the provided explanation) to compensate for inaccuracies in the AI-generated information ((\(1 - \alpha\))) and to deliver an overall expected value exceeding the guaranteed profit \(v\). If the AI delegation condition for FS \eqref{eq:cond_2} is satisfied, it is prioritized over delegation to the data engineer. However, this preference also depends on the accuracy of AI-generated information being sufficiently high compared to the profit \(v\). Specifically, the ''FS feasibility condition'' must hold:

\begin{equation}
\begin{aligned}
\alpha^{*}_{FS}  & > v.
\end{aligned}
\label{eq:cond_3}
\end{equation}

Condition \eqref{eq:cond_3} ensures that the expected gain (\(\alpha \cdot (+1)\)) without validation exceeds the guaranteed profit, \(v\). If this condition is not satisfied, it is not justified to use FS, even with perfect validation effectiveness, as the AI's base accuracy would be insufficient to outperform human delegation. The accuracy threshold \eqref{eq:cond_3} for FS is lower than the threshold \eqref{eq:cond_1} for 
\[
\alpha^*_{FS} < \alpha^*_{PS}. \]
This indicates that the inclusion of a validation step in FS broadens the potential applicability of CBA. However, whether FS is advantageous compared with delegation to a data engineer depends on the level of validation effectiveness, as defined in the AI delegation condition for FS \eqref{eq:cond_2}.

The end user, acting as a rational decision-maker, will only increase the AI support level from PS to FS if the probability of successful validation exceeds the probability of successful SQL generation. This requirement is formalized as the ''validation dominance condition'':

\begin{equation}
\begin{aligned}
E_{FS}(\alpha, \beta) & > E_{PS}(\beta) \Leftrightarrow \beta^{**}  > \alpha.
\end{aligned}
\label{eq:cond_4}
\end{equation}

All \(\beta\) values that satisfy \eqref{eq:cond_4} make FS more advantageous than PS. It is important to note that all the conditions described so far  depend on the values of the potential gain (\(+1\)) and loss (\(-1\)). If the value of the loss outweighs that of the gain, the threshold for \(\beta\) in the validation dominance condition shifts, allowing a less stringent validation standard for FS to outperform PS.

\begin{figure*}[h]
	\begin{center}
		    \includegraphics[width=0.6\textwidth]{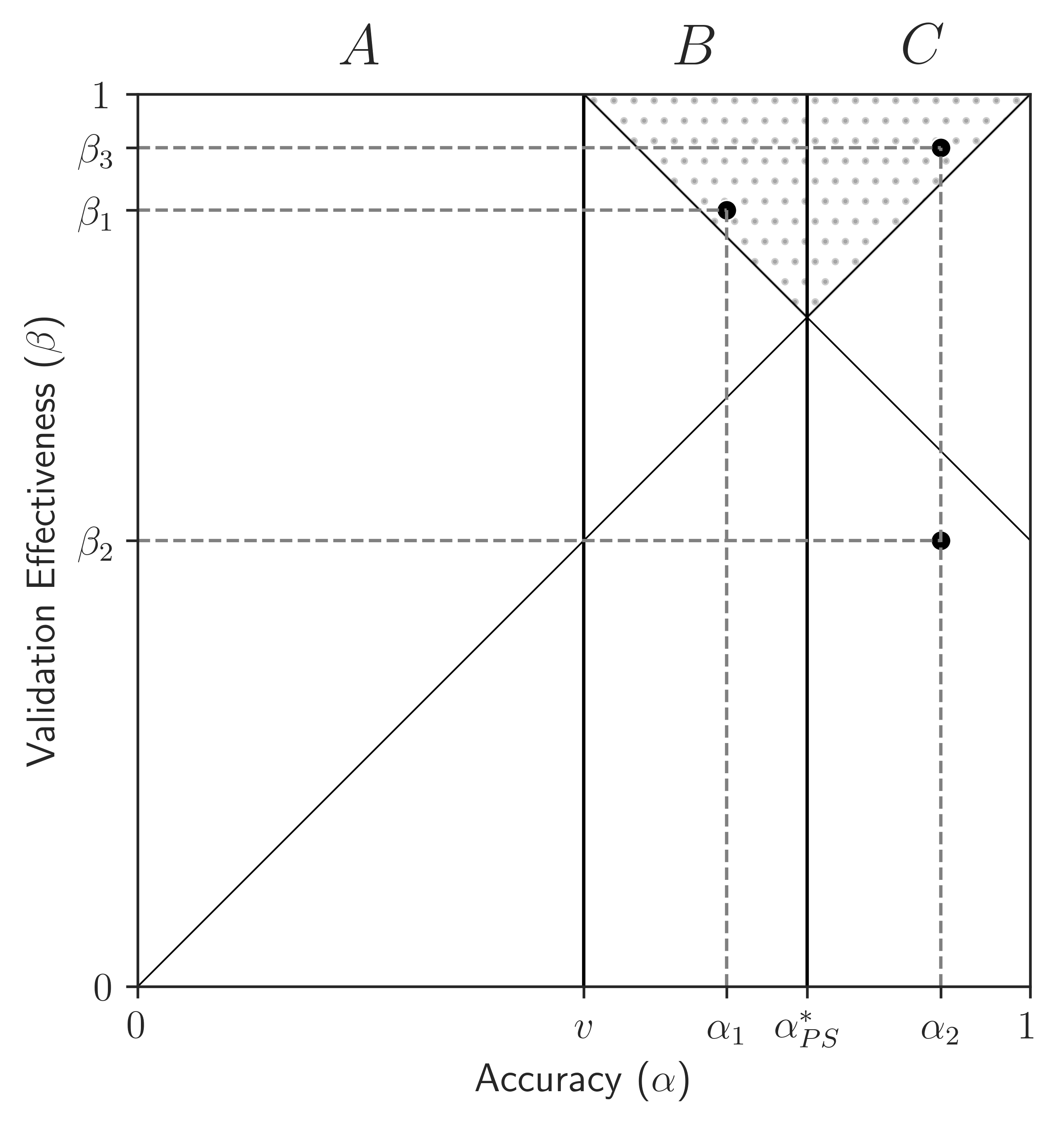} 
    \caption{Relationship between accuracy (\(\alpha\)) and validation effectiveness (\(\beta\)): conditions for effective AI delegation.}
    \label{fig:result}
	\end{center}
\end{figure*}

Fig. \ref{fig:result} illustrates the relationship between accuracy, validation effectiveness and the profit achievable through delegation to the data engineer (all values are measured in monetary units). The dotted area highlights the range of \(\alpha\) and \(\beta\) values where FS outperforms PS and the data engineer, as all three conditions regarding AI delegation \eqref{eq:cond_2}, feasibility of FS \eqref{eq:cond_3}, and dominance of validation \eqref{eq:cond_4} are satisfied. Outside this area, FS does not yield sufficient expected value to justify its use. The figure also identifies the threshold \( \alpha^*_{PS}\) which is determined by the AI delegation condition for PS \eqref{eq:cond_1}.

The analysis reveals three distinct regions based on the interplay between accuracy, validation effectiveness, and the profit achievable through delegation. 

In region \(A\), the FS feasibility condition \eqref{eq:cond_3} is not met, because the accuracy of information generation is less than or equal to the profit achievable through the data engineer (\(\alpha \leq v\)). 
In this region, neither PS nor FS offer sufficient expected values. PS fails owing to its low accuracy, and FS is unable to compensate for this limitation, even when validation effectiveness would be perfect. Delegation to the data engineer remains the only viable option for retrieving the KPI. In this region, CBA is not utilized because it fails to match the effectiveness of the data engineer, rendering it unsuccessful. Furthermore, if the data engineer provides the KPI with significant delay, rendering it irrelevant for the end user due to its diminished value for timely decision-making (\(v\) approaching zero), this not only signifies the failure of CBA but also highlights a broader limitation of business analytics in fulfilling its fundamental objective of delivering timely and decision-relevant information.

Moreover, the requirements for accuracy in PS and FS, as well as for validation effectiveness in FS, increase as \(v\) rises. A reduction in urgency causes a rightward shift of \(v\), \(\alpha^*_{PS}\), and the delegation threshold condition for FS \eqref{eq:cond_2}. In other words, the less urgent the KPI (making it less problematic to wait for the data engineer), the greater the performance demands on CBA to generate the correct SQL query and support the reliable assessment of its correctness.

In Region \(B\), FS feasibility condition \eqref{eq:cond_3} is met (\(\alpha > v\)), but the AI delegation condition for PS \eqref{eq:cond_1} is not satisfied (\(\alpha \leq \alpha^*_{PS}\)). In this region, PS is less effective than delegating to the data engineer.
However, FS becomes a viable option if the validation effectiveness satisfies the AI delegation condition for FS \eqref{eq:cond_2}. Under this condition, validation introduces a ''boosting effect'' by identifying and mitigating errors in SQL queries, ensuring that flawed KPIs do not influence the decision-making process. This boosting effect enables FS to enhance the overall reliability of the information, compensating for the limitations of lower accuracy. For example, even if the accuracy falls below the threshold required for PS \eqref{eq:cond_1} (e.g., \(\alpha_1\)), a sufficiently high validation effectiveness in FS (e.g., \(\beta_1\)) can still yield an expected value greater than \(v\). While the generation of the SQL query may fail, resulting in incorrect KPI, sufficiently effective validation can still identify these errors, ensuring that incorrect KPIs do not adversely affect the decision process.

In region \(C\), where the AI delegation condition for PS \eqref{eq:cond_1} is met (\(\alpha > \alpha^*_{PS}\)), PS becomes a viable alternative to delegation to the data engineer as it provides sufficient benefit on its own. If the AI delegation condition for FS \eqref{eq:cond_2} and the validation dominance condition \eqref{eq:cond_4} are not fulfilled (e.g., at \(\alpha_2\) and \(\beta_2\)), FS has a ''devastating effect'' because the validation process undermines the reliability of the SQL queries generated. Not only are erroneous SQL queries less effectively identified, but insufficient validation also leads to the unwarranted rejection of correct SQL queries. Consequently, FS is not only less effective than PS but also inferior to delegation to the data engineer. Therefore, AI-based assistance for KPI generation with AI-supported validation is not recommended. This underscores the inherent risk of relying on user-based validation without robust mechanisms to support the correctly evaluation of AI-generated SQL queries. In cases where expertise is low or explanations are unclear, FS may inadvertently introduce errors or misjudgments.  

FS becomes favorable when, in region \(C\), conditions \eqref{eq:cond_2} and \eqref{eq:cond_4} are fulfilled (e.g., at \(\beta_2\) and \(\alpha_3\)). In such cases, the validation process enhances the reliability of the generated KPI and maximizes the benefits of FS. As the accuracy increases, the requirements for validation effectiveness must also increase to ensure that FS remains beneficial. A particular challenge arises when the accuracy is very high; in such cases, user-based validation must become even more effective to provide added value over PS. This places significant demands on validation mechanisms, particularly in terms of users' ability to evaluate and assess complex SQL queries based on the given explanations.

\section{Conclusion}

This study analyzes CBA, an emerging approach with considerable potential to address the skill gaps inherent in traditional self-service analytics. Through the use of LLMs, CBA facilitates natural language interactions, enabling end users to independently perform tasks such as data retrieval, preparation, and insight generation. In this context, CBA represents a novel approach to task delegation in business analytics, shifting the process of information generation from human experts to AI, thereby making advanced analytics more timely and accessible to a broader range of organizational roles. This study focuses on Text-to-SQL, a  semantic parsing technology that has seen transformative advancements through the application of LLMs, enhancing its ability to translate natural language requests into structured SQL statements. 

The theoretical models developed in this study examine the conditions under delegation to AI, through PS and FS, outperforms the delegation of tasks to human experts. PS is effective when the accuracy (\(\alpha\)) exceeds a specified threshold, as outlined in condition \eqref{eq:cond_1}. FS, however, is only beneficial when three conditions are met:  AI delegation condition \eqref{eq:cond_2}, FS feasibility condition \eqref{eq:cond_3}, and validation dominance condition \eqref{eq:cond_4}.

This study highlights the importance of validation effectiveness (\(\beta\)) in the context of FS. FS can exert a ''boosting effect'' on information generation, particularly when the validation is robust. Even with low accuracy ( \(\alpha\)), a positive benefit can be derived from the delegation process by preventing incorrect information from being incorporated into operational decision-making. However, the reliance on user-based validation introduces significant uncertainties, especially when users lack the necessary expertise or when SQL queries and data models are highly complex. These challenges can have a ''devastating effect'' on the quality of information generation, potentially making FS less effective than PS and, in the worst-case scenario, even less effective than delegation to a data engineer. This underscores the need for robust validation mechanisms that support in correctly evaluating AI-generated information. Without such mechanisms, the advantages of CBA -- particularly when the accuracy (\(\alpha\) ) of translating natural language requests into correct information is high -- could be significantly diminished.

A promising avenue for future research is the development of hybrid approaches that combine existing techniques (e.g., decomposition, visualization and dialogue) to support code comprehension with automated validation techniques. Future studies could explore the implementation of automated methods for user-independent assessment of the reliability of the generated KPIs. Potential approaches might include statistical methods, such as confidence intervals, to evaluate the correctness of KPIs, and cross-validation with previously validated reports and KPIs. Integrating such methods could mitigate the risk of misclassification owing to a lack of expertise, thereby enhancing the robustness of CBA.

The theoretical models presented in this study are based on several simplifying assumptions. First, the gain from making the correct decision is fixed at \(+1\) and the loss from making the wrong decision is set to \(-1\). Although these values determine the derived conditions, they may not fully capture the variability observed in real-world scenarios. Furthermore, the analysis adopts a static perspective, assuming a single execution cycle for KPI creation and validation, thereby overlooking the iterative nature of CBA. The models also disregard the costs associated with using LLMs, data storage, and processing, potentially leading to overestimation of the economic feasibility of CBA. Additionally, they assume perfect data quality, which may not align with practical conditions. Extending these models to incorporate dynamic, cost-sensitive approaches to information generation and account for data quality considerations could provide deeper and more realistic understanding.

\bibliographystyle{plain}
\bibliography{references_cba}

\begin{thebibliography}{10}

\bibitem{Alpar2016}
Paul Alpar and Michael Schulz.
\newblock Self-service business intelligence.
\newblock {\em Business \& Information Systems Engineering}, 58:151--155, 02
  2016.

\bibitem{Alparslan2024_b}
Adem Alparslan and Denis Chernenko.
\newblock Ai-powered opportunities for business analytics – self-service
  analytics goes conversational.
\newblock {\em BI-Spektrum}, pages 26--30, 2023.

\bibitem{Alparslan2023}
Adem Alparslan and Torben Hugens.
\newblock Business analytics im mittelstand.
\newblock {\em Wirtschaftsinformatik \& Management}, 15(2), 2023.

\bibitem{Askari2024}
Arian Askari, Christian Poelitz, and Xinye Tang.
\newblock Magic: Generating self-correction guideline for in-context
  text-to-sql, 2024.

\bibitem{BanHani2019}
Imad Bani-Hani, Olgerta Tona, and Sven Carlsson.
\newblock Modes of engagement in ssba: a service dominant logic perspective.
\newblock In {\em Proceedings of Americas Conference on Information Systems.
  Association for Information Systems}, 2019.

\bibitem{Beck2024}
Tobi Beck.
\newblock How to simplify sql with text-to-sql technology.
\newblock
  \url{https://www.eckerson.com/articles/how-to-simplify-sql-with-text-to-sql-technology},
  2024.
\newblock Accessed: 2024-11-15.

\bibitem{Brown2020}
Tom~B. Brown, Benjamin Mann, Nick Ryder, Melanie Subbiah, Jared Kaplan,
  Prafulla Dhariwal, Arvind Neelakantan, Pranav Shyam, Girish Sastry, Amanda
  Askell, Sandhini Agarwal, Ariel Herbert-Voss, Gretchen Krueger, Tom Henighan,
  Rewon Child, Aditya Ramesh, Daniel~M. Ziegler, Jeffrey Wu, Clemens Winter,
  Christopher Hesse, Mark Chen, Eric Sigler, Mateusz Litwin, Scott Gray,
  Benjamin Chess, Jack Clark, Christopher Berner, Sam McCandlish, Alec Radford,
  Ilya Sutskever, and Dario Amodei.
\newblock Language models are few-shot learners, 2020.

\bibitem{Chan2005}
Hock~Chuan Chan, Hock-Hai Teo, and XH~Zeng.
\newblock An evaluation of novice end-user computing performance: Data
  modeling, query writing, and comprehension.
\newblock {\em Journal of the American Society for Information Science and
  Technology}, 56(8):843--853, 2005.

\bibitem{Chaudhuri1997}
Surajit Chaudhuri and Umeshwar Dayal.
\newblock An overview of data warehousing and olap technology.
\newblock {\em SIGMOD Rec.}, 26(1):65--74, 1997.

\bibitem{Chen2012}
Hsinchun Chen, Roger H.~L. Chiang, and Veda~C. Storey.
\newblock Business intelligence and analytics: From big data to big impact.
\newblock {\em MIS Quarterly}, 36(4):1165--1188, 2012.

\bibitem{Chen2023_window}
Shouyuan Chen, Sherman Wong, Liangjian Chen, and Yuandong Tian.
\newblock Extending context window of large language models via positional
  interpolation, 2023.

\bibitem{Chen2024}
Sibei Chen, Hanbing Liu, Waiting Jin, Xiangyu Sun, Xiaoyao Feng, Ju~Fan,
  Xiaoyong Du, and Nan Tang.
\newblock Chatpipe: Orchestrating data preparation pipelines by optimizing
  human-chatgpt interactions.
\newblock In {\em Companion of the 2024 International Conference on Management
  of Data}, pages 484--487, 2024.

\bibitem{Codd1974}
E.~F. Codd.
\newblock Seven steps to rendezvous with the casual user.
\newblock In {\em IFIP Working Conference Data Base Management}, 1974.

\bibitem{Codd1993}
Edgar~F. Codd, Sally~B. Codd, and Clynch~T. Salley.
\newblock Providing {OLAP} to user-analysts: An {IT}-mandate.
\newblock {\em Codd and Associates}, 1993.

\bibitem{DARPA2016}
DARPA.
\newblock Explainable artificial intelligence (xai).
\newblock Technical Report DARPA-BAA-16-53, Defense Advanced Research Projects
  Agency, 2016.

\bibitem{Dejeu2024}
Emily~Barrow DeJeu.
\newblock Using generative ai to facilitate data analysis and visualization: A
  case study of olympic athletes.
\newblock {\em Journal of Business and Technical Communication}, page
  10506519241239923, 2024.

\bibitem{Delen2013}
Dursun Delen and Haluk Demirkan.
\newblock Data, information and analytics as services.
\newblock {\em Decision Support Systems}, 55:359--363, 04 2013.

\bibitem{Delen2018}
Dursun Delen and Sudha Ram.
\newblock Research challenges and opportunities in business analytics.
\newblock {\em Journal of Business Analytics}, 1(1):2--12, 2018.

\bibitem{Devlin2019}
Jacob Devlin, Ming-Wei Chang, Kenton Lee, and Kristina Toutanova.
\newblock Bert: Pre-training of deep bidirectional transformers for language
  understanding.
\newblock In {\em North American Chapter of the Association for Computational
  Linguistics}, 2019.

\bibitem{Elgohary2020}
Ahmed Elgohary, Saghar Hosseini, and Ahmed~Hassan Awadallah.
\newblock Speak to your parser: Interactive text-to-sql with natural language
  feedback, 2020.

\bibitem{Eusebius2024}
Nitin Eusebius, Arghya Banerjee, and Randy DeFauw.
\newblock Generating value from enterprise data: Best practices for text2sql
  and generative ai.
\newblock
  \url{https://aws.amazon.com/de/blogs/machine-learning/generating-value-from-enterprise-data-best-practices-for-text2sql-
  \\and-generative-ai/}.
\newblock Accessed: 2024-11-15.

\bibitem{Floratou2024}
Avrilia Floratou, Fotis Psallidas, Fuheng Zhao, Shaleen Deep, Gunther
  Hagleither, Wangda Tan, Joyce Cahoon, Rana Alotaibi, Jordan Henkel, Abhik
  Singla, Alex~Van Grootel, Brandon Chow, Kai Deng, Katherine Lin, Marcos
  Campos, K.~Venkatesh Emani, Vivek Pandit, Victor Shnayder, Wenjing Wang, and
  Carlo Curino.
\newblock Nl2sql is a solved problem... not!
\newblock In {\em CIDR}, 2024.

\bibitem{Glinz2015}
Martin Glinz and Samuel~A. Fricker.
\newblock On shared understanding in software engineering: an essay.
\newblock {\em Computer Science - Research and Development}, 30(3):363--376,
  2015.

\bibitem{Gunning2019}
David Gunning and David Aha.
\newblock Darpa’s explainable artificial intelligence (xai) program.
\newblock {\em AI magazine}, 40(2):44--58, 2019.

\bibitem{Guo_RAG_2023}
Chunxi Guo, Zhiliang Tian, Jintao Tang, Shasha Li, Zhihua Wen, Kaixuan Wang,
  and Ting Wang.
\newblock Retrieval-augmented gpt-3.5-based text-to-sql framework
  with sample-aware prompting and dynamic revision chain.
\newblock In Biao Luo, Long Cheng, Zheng-Guang Wu, Hongyi Li, and Chaojie Li,
  editors, {\em Neural Information Processing}, pages 341--356. Springer Nature
  Singapore, 2024.

\bibitem{Gur2018}
Izzeddin G{\"u}r, Semih Yavuz, Yu~Su, and Xifeng Yan.
\newblock Dialsql: Dialogue based structured query generation.
\newblock In {\em Proceedings of the 56th Annual Meeting of the Association for
  Computational Linguistics (Volume 1: Long Papers)}, pages 1339--1349, 2018.

\bibitem{Gurbaxani1990}
Vijay Gurbaxani and Chris~F. Kemerer.
\newblock An agency theory view of management of end-user computing.
\newblock In {\em ICIS 1990 Proceedings}, 1990.

\bibitem{Hirschheim1995}
Rudy Hirschheim, Heinz~K Klein, and Kalle Lyytinen.
\newblock {\em Information systems development and data modeling: conceptual
  and philosophical foundations}, volume~9.
\newblock Cambridge University Press, 1995.

\bibitem{Holsapple2014}
Clyde~W. Holsapple, Anita Lee-Post, and Ramakrishnan Pakath.
\newblock A unified foundation for business analytics.
\newblock {\em Decision Support Systems}, 64:130--141, 2014.

\bibitem{Imhoff2011}
Claudia Imhoff and Colin White.
\newblock Self-service business intelligence: Empowering users to generate
  insights.
\newblock {\em TDWI Best practices report}, 2011.

\bibitem{Inmon2015}
W.H. Inmon and Daniel Linstedt.
\newblock {\em Data Architecture: a Primer for the Data Scientist}.
\newblock Morgan Kaufmann, 2015.

\bibitem{Iriarte2020}
Carmen Iriarte and Sussy Bayona.
\newblock It projects success factors: a literature review.
\newblock {\em International Journal of Information Systems and Project
  Management}, 8:15, 2020.

\bibitem{Jeong2023}
Geunyeong Jeong, Mirae Han, Seulgi Kim, Yejin Lee, Joosang Lee, Seongsik Park,
  and Harksoo Kim.
\newblock Improving text-to-sql with a hybrid decoding method.
\newblock {\em Entropy}, 25(3), 2023.

\bibitem{Katsogiannis2023}
George Katsogiannis-Meimarakis and Georgia Koutrika.
\newblock A survey on deep learning approaches for text-to-sql.
\newblock {\em The VLDB Journal}, 32(4):905--936, 2023.

\bibitem{Koch2023}
Christian Koch, Markus Stadi, and Lukas Berle.
\newblock From data engineering to prompt engineering: Solving data preparation
  tasks with chatgpt, 2023.

\bibitem{Kokkalis2012}
Andreas Kokkalis, Panagiotis Vagenas, Alexandros Zervakis, Alkis Simitsis,
  Georgia Koutrika, and Yannis Ioannidis.
\newblock Logos: a system for translating queries into narratives.
\newblock In {\em Proceedings of the 2012 ACM SIGMOD International Conference
  on Management of Data}, SIGMOD '12, page 673–676. Association for Computing
  Machinery, 2012.

\bibitem{Langer2021}
Markus Langer, Daniel Oster, Timo Speith, Holger Hermanns, Lena Kästner, Eva
  Schmidt, Andreas Sesing, and Kevin Baum.
\newblock What do we want from explainable artificial intelligence (xai)? – a
  stakeholder perspective on xai and a conceptual model guiding
  interdisciplinary xai research.
\newblock {\em Artificial Intelligence}, 296:103473, 2021.

\bibitem{Lennerholt2021}
Christian Lennerholt, Joeri~Van Laere, and Eva Söderström.
\newblock User-related challenges of self-service business intelligence.
\newblock {\em Information Systems Management}, 38(4):309--323, 2021.

\bibitem{Leventidis2020}
Aristotelis Leventidis, Jiahui Zhang, Cody Dunne, Wolfgang Gatterbauer, H.V.
  Jagadish, and Mirek Riedewald.
\newblock Queryvis: Logic-based diagrams help users understand complicated sql
  queries faster.
\newblock In {\em Proceedings of the 2020 ACM SIGMOD International Conference
  on Management of Data}, SIGMOD '20, page 2303–2318. Association for
  Computing Machinery, 2020.

\bibitem{Li2023}
Huaxia Li, Haoyun Gao, Chengzhang Wu, and Miklos~A. Vasarhelyi.
\newblock Extracting financial data from unstructured sources: Leveraging large
  language models, 2023.

\bibitem{Li_BIRD_2023}
Jinyang Li, Binyuan Hui, GE~QU, Jiaxi Yang, Binhua Li, Bowen Li, Bailin Wang,
  Bowen Qin, Ruiying Geng, Nan Huo, Xuanhe Zhou, Chenhao Ma, Guoliang Li, Kevin
  Chang, Fei Huang, Reynold Cheng, and Yongbin Li.
\newblock Can {LLM} already serve as a database interface? a {BI}g bench for
  large-scale database grounded text-to-{SQL}s.
\newblock In {\em Thirty-seventh Conference on Neural Information Processing
  Systems Datasets and Benchmarks Track}, 2023.

\bibitem{Li2024_NLDB}
Yunyao Li, Dragomir Radev, and Davood Rafiei.
\newblock {\em Natural language interfaces to databases}.
\newblock Springer, 2024.

\bibitem{Lin2024}
Zichao Lin, Shuyan Guan, Wending Zhang, Huiyan Zhang, Yugang Li, and Huaping
  Zhang.
\newblock Towards trustworthy llms: a review on debiasing and dehallucinating
  in large language models.
\newblock {\em Artificial Intelligence Review}, 57(9):243, 2024.

\bibitem{Liu2023}
Pengfei Liu, Weizhe Yuan, Jinlan Fu, Zhengbao Jiang, Hiroaki Hayashi, and
  Graham Neubig.
\newblock Pre-train, prompt, and predict: A systematic survey of prompting
  methods in natural language processing.
\newblock {\em ACM Comput. Surv.}, 55(9), January 2023.

\bibitem{Llave2018}
Marilex~Rea Llave.
\newblock Data lakes in business intelligence: reporting from the trenches.
\newblock {\em Procedia Computer Science}, 138:516--524, 2018.

\bibitem{Maddigan2023}
Paula Maddigan and Teo Susnjak.
\newblock Chat2vis: generating data visualizations via natural language using
  chatgpt, codex and gpt-3 large language models.
\newblock {\em Ieee Access}, 11:45181--45193, 2023.

\bibitem{Michalczyk2020}
Sven Michalczyk, Mario Nadj, Darius Azarfar, Alexander Maedche, and Christoph
  Groeger.
\newblock A state-of-the-art overview and future research avenues of
  self-service business intelligence and analytics.
\newblock In {\em ECIS 2020 Proceedings -- Twenty-Eighth European Conference on
  Information Systems, Marrakesh, Marokko, June 15 - 17, 2020}, 2020.

\bibitem{Miedema2021}
Daphne Miedema and George Fletcher.
\newblock Sqlvis: Visual query representations for supporting sql learners.
\newblock In {\em 2021 IEEE Symposium on Visual Languages and Human-Centric
  Computing (VL/HCC)}, pages 1--9, 2021.

\bibitem{Minh2022}
Dang Minh, H.~Xiang Wang, Y.~Fen Li, and Tan~N. Nguyen.
\newblock Explainable artificial intelligence: a comprehensive review.
\newblock {\em Artificial Intelligence Review}, 55(5):3503--3568, 2022.

\bibitem{Monteiro2023}
Walbert~Cunha Monteiro, Diego~Hortêncio Dos~Santos, Thiago Augusto~Soares
  De~Sousa, Vinicius~Favacho Queiroz, Tiago Davi~Oliveira De~Araújo, and
  Bianchi~Serique Meiguins.
\newblock Workload evaluation to create data visualization using chatgpt.
\newblock In {\em 2023 27th International Conference Information Visualisation
  (IV)}, pages 136--141, 2023.

\bibitem{Mori1995}
Angelo~Rossi Mori.
\newblock Coding systems and controlled vocabularies for hospital information
  systems.
\newblock {\em International Journal of Bio-Medical Computing}, 39(1):93--98,
  1995.

\bibitem{Munoz2020}
Marvin Munoz~Baron, Marvin Wyrich, and Stefan Wagner.
\newblock An empirical validation of cognitive complexity as a measure of
  source code understandability.
\newblock In {\em Proceedings of the 14th ACM / IEEE International Symposium on
  Empirical Software Engineering and Measurement (ESEM)}, page 1–12. ACM,
  2020.

\bibitem{Murakawa2011}
Takehiko Murakawa and Masaru Nakagawa.
\newblock Comprehension support of sql statement using double-tree structure.
\newblock In {\em CSEDU 2011 - Proceedings of the 3rd International Conference
  on Computer Supported Education}, volume~1, pages 318--323, 01 2011.

\bibitem{Musazade2024}
Nurlan Musazade, J{\'o}zsef Mezei, and Xiaolu Wang.
\newblock Exploring the performance of large language models for data analysis
  tasks through the crisp-dm framework.
\newblock In {\'A}lvaro Rocha, Hojjat Adeli, Gintautas Dzemyda, Fernando
  Moreira, and Aneta Poniszewska-Mara{\'{n}}da, editors, {\em Good Practices
  and New Perspectives in Information Systems and Technologies}, pages 56--65.
  Springer Nature Switzerland, 2024.

\bibitem{Nakatsu2006}
Robbie~T. Nakatsu.
\newblock {\em Explanatory Power of Intelligent Systems}, pages 123--143.
\newblock Springer London, 2006.

\bibitem{Nambiar2022}
Athira Nambiar and Divyansh Mundra.
\newblock An overview of data warehouse and data lake in modern enterprise data
  management.
\newblock {\em Big Data and Cognitive Computing}, 6(4), 2022.

\bibitem{Arpit2021}
Arpit Narechania, Adam Fourney, Bongshin Lee, and Gonzalo Ramos.
\newblock Diy: Assessing the correctness of natural language to sql systems.
\newblock In {\em Proceedings of the 26th International Conference on
  Intelligent User Interfaces}, IUI '21, page 597–607. Association for
  Computing Machinery, 2021.

\bibitem{Nasseri2023}
Mehran Nasseri, Patrick Brandtner, Robert Zimmermann, Taha Falatouri, Farzaneh
  Darbanian, and Tobechi Obinwanne.
\newblock Applications of large language models (llms) in business
  analytics--exemplary use cases in data preparation tasks.
\newblock In {\em International Conference on Human-Computer Interaction},
  pages 182--198. Springer, 2023.

\bibitem{Nelson1989}
R.~Ryan Nelson, editor.
\newblock {\em End-user computing: Concepts, issues, and applications}.
\newblock John Wiley \& Sons, Inc., 1989.

\bibitem{Zhang2023__}
Zheng Ning, Zheng Zhang, Tianyi Sun, Yuan Tian, Tianyi Zhang, and Toby Jia-Jun
  Li.
\newblock An empirical study of model errors and user error discovery and
  repair strategies in natural language database queries.
\newblock In {\em Proceedings of the 28th International Conference on
  Intelligent User Interfaces}, pages 633--649, 2023.

\bibitem{Palys2023}
Marcin Pa{\l}ys and Andrzej Pa{\l}ys.
\newblock Benefits and challenges of self-service business intelligence
  implementation.
\newblock {\em Procedia Computer Science}, 225:795--803, 2023.
\newblock 27th International Conference on Knowledge Based and Intelligent
  Information and Engineering Sytems (KES 2023).

\bibitem{Passlick2023}
Jens Passlick, Lukas Gr{\"u}tzner, Michael Schulz, and Michael~H. Breitner.
\newblock Self-service business intelligence and analytics application
  scenarios: A taxonomy for differentiation.
\newblock {\em Information Systems and e-Business Management}, 21(1):159--191,
  2023.

\bibitem{Perkovic2024}
Gabrijela Perković, Antun Drobnjak, and Ivica Botički.
\newblock Hallucinations in llms: Understanding and addressing challenges.
\newblock In {\em 2024 47th MIPRO ICT and Electronics Convention (MIPRO)},
  pages 2084--2088, 2024.

\bibitem{Pourreza2024}
Mohammadreza Pourreza and Davood Rafiei.
\newblock Din-sql: decomposed in-context learning of text-to-sql with
  self-correction.
\newblock In {\em Proceedings of the 37th International Conference on Neural
  Information Processing Systems}, NIPS '23. Curran Associates Inc., 2024.

\bibitem{Prabhu2019}
CSR Prabhu, Aneesh~Sreevallabh Chivukula, Aditya Mogadala, Rohit Ghosh,
  LM~Jenila Livingston, CSR Prabhu, Aneesh~Sreevallabh Chivukula, Aditya
  Mogadala, Rohit Ghosh, and LM~Jenila Livingston.
\newblock {\em Big data analytics}.
\newblock Springer, 2019.

\bibitem{Raffel2020}
Colin Raffel, Noam Shazeer, Adam Roberts, Katherine Lee, Sharan Narang, Michael
  Matena, Yanqi Zhou, Wei Li, and Peter~J. Liu.
\newblock Exploring the limits of transfer learning with a unified text-to-text
  transformer.
\newblock {\em Journal of Machine Learning Research}, 21(140):1--67, 2020.

\bibitem{Rudin2019}
Cynthia Rudin.
\newblock Stop explaining black box machine learning models for high stakes
  decisions and use interpretable models instead.
\newblock {\em Nature Machine Intelligence}, 1(5):206--215, 2019.

\bibitem{Schmarzo2023}
Bill Schmarzo.
\newblock {\em AI \& Data Literacy: Empowering Citizens of Data Science}.
\newblock Packt Publishing Ltd, 2023.

\bibitem{Sharma2023}
Ankita Sharma, Xuanmao Li, Hong Guan, Guoxin Sun, Liang Zhang, Lanjun Wang,
  Kesheng Wu, Lei Cao, Erkang Zhu, Alexander Sim, et~al.
\newblock Automatic data transformation using large language model-an
  experimental study on building energy data.
\newblock In {\em 2023 IEEE International Conference on Big Data (BigData)},
  pages 1824--1834. IEEE, 2023.

\bibitem{Stockl2024}
Andreas St{\"o}ckl.
\newblock Information visualization with chatgpt.
\newblock In Boris Kovalerchuk, Kawa Nazemi, R{\u{a}}zvan Andonie, Nuno Datia,
  and Ebad Banissi, editors, {\em Artificial Intelligence and Visualization:
  Advancing Visual Knowledge Discovery}, pages 469--485. Springer Nature
  Switzerland, 2024.

\bibitem{Stodder2015}
D.~Stodder.
\newblock Visual analytics for making smarter decisions faster.
\newblock {\em TDWI Best practices report}, 2015.

\bibitem{Arasteh2024}
Soroosh Tayebi~Arasteh, Tianyu Han, Mahshad Lotfinia, Christiane Kuhl,
  Jakob~Nikolas Kather, Daniel Truhn, and Sven Nebelung.
\newblock Large language models streamline automated machine learning for
  clinical studies.
\newblock {\em Nature Communications}, 15(1):1603, 2024.

\bibitem{Thapa2024}
Astha Thapa and Rajvardhan Patil.
\newblock Chatgpt based chatbot application.
\newblock In {\em SoutheastCon 2024}, pages 157--164, 2024.

\bibitem{Thorpe2024}
Dayton~G. Thorpe, Andrew~J. Duberstein, and Ian~A. Kinsey.
\newblock Dubo-sql: Diverse retrieval-augmented generation and fine tuning for
  text-to-sql, 2024.

\bibitem{Tian2024}
Yuan Tian, Jonathan~K. Kummerfeld, Toby Jia-Jun Li, and Tianyi Zhang.
\newblock Sqlucid: Grounding natural language database queries with interactive
  explanations, 2024.

\bibitem{Tomova2023}
Mihaela Tomova, Martin Hofmann, Constantin H{\"u}tterer, and Patrick M{\"a}der.
\newblock Assessing the utility of text-to-sql approaches for satisfying
  software developer information needs.
\newblock {\em Empirical Software Engineering}, 29(1):15, 2023.

\bibitem{Tran2024}
Quoc-Bao-Huy Tran, Aagha~Abdul Waheed, and Sun-Tae Chung.
\newblock Robust text-to-cypher using combination of bert, graphsage, and
  transformer (cobgt) model.
\newblock {\em Applied Sciences}, 14(17), 2024.

\bibitem{Tunney2024}
Richard Tunney.
\newblock {\em A Primer of Judgment and Decision Making}.
\newblock Springer, 2024.

\bibitem{Vaswani2017}
Ashish Vaswani, Noam Shazeer, Niki Parmar, Jakob Uszkoreit, Llion Jones,
  Aidan~N. Gomez, \L{}ukasz Kaiser, and Illia Polosukhin.
\newblock Attention is all you need.
\newblock In {\em Proceedings of the 31st International Conference on Neural
  Information Processing Systems}, NIPS'17, page 6000–6010. Curran Associates
  Inc., 2017.

\bibitem{Vercellis2009}
Carlo Vercellis.
\newblock {\em Business Intelligence: Data Mining and Optimization for Decision
  Making}.
\newblock Wiley, 1st edition, 2009.

\bibitem{Neumann2007}
John Von~Neumann and Oskar Morgenstern.
\newblock Theory of games and economic behavior: 60th anniversary commemorative
  edition.
\newblock In {\em Theory of games and economic behavior}. Princeton university
  press, 2007.

\bibitem{Vujosevic2018}
Du{\v s}an Vujosevic, Ivana Kovacevic, and Milena Vujosevic-Janicic.
\newblock The learnability of the dimensional view of data and what to do with
  it.
\newblock {\em Aslib Journal of Information Management}, 71(1):38--53, 2018.

\bibitem{Wallis1982}
Jerold~W Wallis and Edward~H Shortliffe.
\newblock Explanatory power for medical expert systems: studies in the
  representation of causal relationships for clinical consultations.
\newblock {\em Methods of Information in Medicine}, 21(03):127--136, 1982.

\bibitem{Xiaxia2021}
Xiaxia Wang, Sai Wu, Lidan Shou, and Ke~Chen.
\newblock An interactive nl2sql approach with reuse strategy.
\newblock In {\em Database Systems for Advanced Applications: 26th
  International Conference, DASFAA 2021, Taipei, Taiwan, April 11--14, 2021,
  Proceedings, Part II 26}, pages 280--288. Springer, 2021.

\bibitem{Warren1982}
David~H.D. Warren and Fernando~C.N. Pereira.
\newblock An efficient easily adaptable system for interpreting natural
  language queries.
\newblock {\em American Journal of Computational Linguistics}, 8(3-4):110--122,
  1982.

\bibitem{Watson2019}
Hugh Watson.
\newblock Update tutorial: Big data analytics: Concepts, technology, and
  applications.
\newblock {\em Communications of the Association for Information Systems},
  44:364--379, 01 2019.

\bibitem{Westbrook1992}
Jack~H Westbrook and Walter Grattidge.
\newblock Terminology standards for materials databases.
\newblock {\em Reference}, 15:15, 1992.

\bibitem{Wu2024}
Yang Wu, Yao Wan, Hongyu Zhang, Yulei Sui, Wucai Wei, Wei Zhao, Guandong Xu,
  and Hai Jin.
\newblock Automated data visualization from natural language via large language
  models: An exploratory study.
\newblock {\em Proceedings of the ACM on Management of Data}, 2(3):1--28, May
  2024.

\bibitem{Yao2019}
Ziyu Yao, Yu~Su, Huan Sun, and Wen-tau Yih.
\newblock Model-based interactive semantic parsing: A unified formulation and a
  text-to-sql case study.
\newblock In {\em 2019 Conference on Empirical Methods in Natural Language
  Processing (EMNLP'19)}, 2019.

\bibitem{yao2019model}
Ziyu Yao, Yu~Su, Huan Sun, and Wen-tau Yih.
\newblock Model-based interactive semantic parsing: A unified framework and a
  text-to-sql case study.
\newblock {\em arXiv preprint arXiv:1910.05389}, 2019.

\bibitem{Yu2019}
Tao Yu, Rui Zhang, Kai Yang, Michihiro Yasunaga, Dongxu Wang, Zifan Li, James
  Ma, Irene Li, Qingning Yao, Shanelle Roman, Zilin Zhang, and Dragomir Radev.
\newblock Spider: A large-scale human-labeled dataset for complex and
  cross-domain semantic parsing and text-to-sql task, 2019.

\bibitem{Zhang2023}
Haochen Zhang, Yuyang Dong, Chuan Xiao, and Masafumi Oyamada.
\newblock Large language models as data preprocessors, 2023.

\bibitem{Zhong2017}
Victor Zhong, Caiming Xiong, and Richard Socher.
\newblock Seq2sql: Generating structured queries from natural language using
  reinforcement learning, 2017.

\end{thebibliography}

\end{document}